\documentclass{article}

\usepackage[preprint]{neurips_2026}

\usepackage[utf8]{inputenc}
\usepackage[T1]{fontenc}
\usepackage{hyperref}
\usepackage{url}
\usepackage{booktabs}
\usepackage{amsmath}
\usepackage{amsfonts}
\usepackage{nicefrac}
\usepackage{microtype}
\usepackage{graphicx}
\usepackage{xcolor}
\usepackage{tikz}
\usepackage{multirow}
\usetikzlibrary{shapes,arrows,positioning,fit,backgrounds}

\title{Compiling Agentic Workflows into LLM Weights:\\Near-Frontier Quality at Two Orders of Magnitude Less Cost}

\author{%
  Simon Dennis \\
  i14, University of Melbourne \\
  \And
  Rivaan Patil \\
  i14 \\
  \And
  Kevin Shabahang \\
  i14 \\
  \And
  Hao Guo \\
  i14 \\
}

\begin{document}

\maketitle

\begin{abstract}
Agent orchestration frameworks have proliferated, collectively exceeding 290{,}000 GitHub stars across LangGraph, CrewAI, Google ADK, OpenAI Agents SDK, Semantic Kernel, Strands, and LlamaIndex. All follow the same pattern: an external orchestrator above the LLM, injecting instructions and routing decisions every turn. Recent work has shown this architecture is dominated for procedural tasks by simply providing the procedure in a frontier model's system prompt~\citep{dennis2026incontext}, at the cost of consuming the context window, requiring a frontier model for every conversation, and exposing proprietary procedures to third-party providers. \textit{Compiling} the procedure into the weights of a small fine-tuned model---creating a \textit{subterranean agent}---should resolve all of these concerns, and prior work (SimpleTOD, FireAct, SynTOD, WorkflowLLM, Agent Lumos) has shown the technique works. Yet developer adoption has overwhelmingly favored orchestration. We identify three perceived barriers and address each empirically across travel booking (14 nodes), Zoom support (14 nodes, product-specific knowledge), and insurance claims (55 nodes, 6 decision hubs).

\textbf{Quality.} On a travel booking task, a controlled same-model comparison isolates the effect of compilation: a 3B compiled model beats the same 3B base model with explicit orchestration on four of five quality metrics ($p < 0.001$). The 3B model reaches only $\sim$82\% of the frontier in-context baseline on graceful handling and naturalness; scaling to 8B on Zoom support and insurance claims closes that gap, achieving 87--98\% of in-context quality and matching a LangGraph orchestrator built on a $\sim$70$\times$ larger frontier model. Compiled models also have lower failure rates than the orchestrator in travel (5.5\% vs.\ 24\%) and insurance (9\% vs.\ 17\%); Zoom failure rates are comparable.

\textbf{Cost.} Compiled models are 128--462$\times$ cheaper per conversation than the in-context baseline, combining a $\sim$65$\times$ per-token reduction from self-hosting with a roughly $2$--$7\times$ token volume reduction. The advantage grows with procedure complexity because the compiled model's prompt is constant-size. Local inference also reduces latency (2.8$\times$ faster in insurance).

\textbf{Flexibility.} The recompile cycle takes 30--50 minutes on production hardware---a CI/CD cycle, not the prohibitively long retraining often assumed.

Each barrier turns out to be smaller than commonly assumed: the quality gap is small (87--98\% of frontier), the cost gap favors compilation by two orders of magnitude (and grows with procedure complexity), and the flexibility gap is a deployment cycle rather than a paradigm shift. Compilation is the natural solution to procedural knowledge that needs to persist beyond a single conversation's context: persistent structure belongs in the weights, transient state belongs in the prompt.
\end{abstract}

\section{Introduction}

Agent orchestration frameworks---LangGraph~\citep{langgraph2024}, CrewAI~\citep{crewai2024}, Google ADK~\citep{googleadk2025}, OpenAI's Agents SDK~\citep{openaiagents2025}, Semantic Kernel~\citep{semantickernel2025}, Strands~\citep{strandsagents2025}, LlamaIndex~\citep{llamaindex2025}---have proliferated, collectively exceeding 290{,}000 GitHub stars. All follow the same pattern: an external orchestrator above the LLM, injecting instructions and routing decisions every turn. \citet{dennis2026incontext} showed that for procedural tasks this architecture is dominated by a simpler alternative---giving the model the entire procedure in its system prompt and letting it self-orchestrate, achieving near-perfect quality (4.53--5.00 on a 5-point scale). However, the in-context approach requires a frontier model for every conversation, inflates token consumption with the procedure embedded in every API call, consumes context window capacity, and exposes proprietary procedures to third-party providers.

\textit{Compiling} the procedure into the weights of a small fine-tuned model---creating what we call a \textit{subterranean agent}---should resolve all of these concerns. The technique is established: SimpleTOD~\citep{hosseini2020simpletod}, FireAct~\citep{chen2023fireact}, SynTOD~\citep{samarinas2024syntod}, Workflow\-LLM~\citep{shao2024workflowllm}, and Agent Lumos~\citep{yin2024lumos} all compile agent capabilities into model weights, and several reach quality competitive with frontier models. Yet developer adoption has overwhelmingly favored orchestration; the compilation papers above remain academic prototypes with $\sim$3{,}000 stars combined---roughly 100$\times$ less community engagement than the orchestration frameworks. \textbf{Why aren't people compiling procedures into weights?}

We identify three perceived barriers and address each empirically across three domains---travel booking (14 nodes), Zoom support (14 nodes, product-specific knowledge), and insurance claims processing (55 nodes, 6 decision hubs)---using $n=200$ scenarios per condition per domain with a LangGraph orchestrator and in-context baseline as controls:

\begin{itemize}
    \item \textbf{Quality} (\S\ref{sec:results}). Will a fine-tuned small model match a frontier model on procedural tasks? An 8B compiled model achieves 87--98\% of in-context frontier quality, competitive with a LangGraph orchestrator using a $\sim$70$\times$ larger frontier model.
    \item \textbf{Cost} (\S\ref{sec:tco}). Is the per-conversation inference cost actually lower once self-hosting is accounted for? Compiled models are 128--462$\times$ cheaper per conversation than the in-context baseline; the advantage grows with procedure complexity because the compiled model's prompt is constant-size.
    \item \textbf{Flexibility} (\S\ref{sec:flexibility}). When the procedure changes, can a compiled model adapt without a prohibitively long retraining effort? The recompile cycle takes 30--50 minutes on production hardware---a CI/CD cycle, not a major retraining job.
\end{itemize}

\section{Compiling Agentic Workflows into Weights}

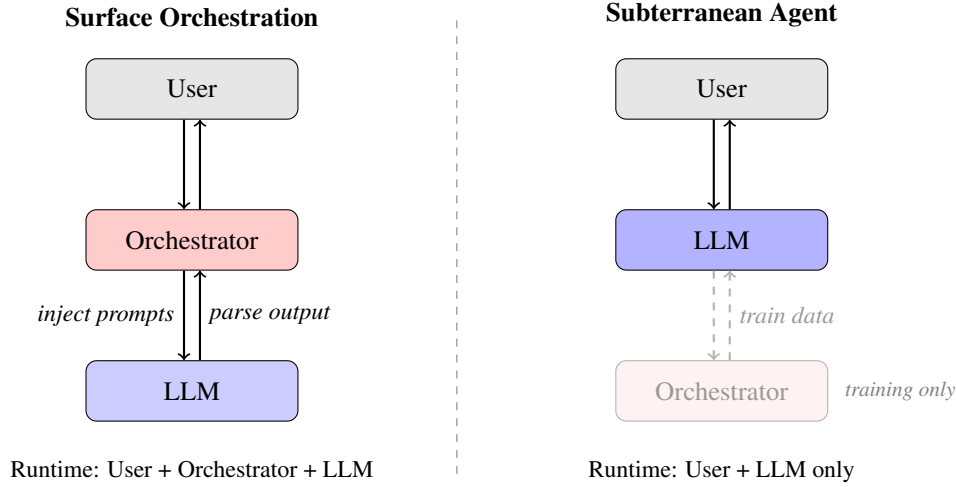
\begin{figure}[t]
\centering
\begin{tikzpicture}[
    box/.style={rectangle, draw, rounded corners, minimum width=2.8cm, minimum height=0.8cm, align=center},
    faded/.style={rectangle, draw, rounded corners, minimum width=2.8cm, minimum height=0.8cm, align=center, draw=black!30, text=black!40},
    arrow/.style={->, thick},
    fadedarrow/.style={->, thick, black!30},
    label/.style={font=\small\itshape}
]

\node[box, fill=gray!20] (user1) at (-3.5, 2.5) {User};
\node[box, fill=red!20] (orch1) at (-3.5, 0.5) {Orchestrator};
\node[box, fill=blue!20] (llm1) at (-3.5, -1.5) {LLM};

\draw[arrow] ([xshift=-3pt]user1.south) -- ([xshift=-3pt]orch1.north);
\draw[arrow] ([xshift=3pt]orch1.north) -- ([xshift=3pt]user1.south);
\draw[arrow] ([xshift=-3pt]orch1.south) -- node[left, label] {inject prompts} ([xshift=-3pt]llm1.north);
\draw[arrow] ([xshift=3pt]llm1.north) -- node[right, label] {parse output} ([xshift=3pt]orch1.south);

\node[above=0.3cm of user1, font=\bfseries] {Surface Orchestration};
\node[below=0.4cm of llm1, font=\small] {Runtime: User + Orchestrator + LLM};

\node[box, fill=gray!20] (user2) at (3.5, 2.5) {User};
\node[box, fill=blue!30] (llm2) at (3.5, 0.5) {LLM};
\node[faded, fill=red!5] (orch2) at (3.5, -1.5) {Orchestrator};

\draw[arrow] ([xshift=-3pt]user2.south) -- ([xshift=-3pt]llm2.north);
\draw[arrow] ([xshift=3pt]llm2.north) -- ([xshift=3pt]user2.south);
\draw[fadedarrow, dashed] ([xshift=-3pt]llm2.south) -- ([xshift=-3pt]orch2.north);
\draw[fadedarrow, dashed] ([xshift=3pt]orch2.north) -- node[right, label, text=black!40] {train data} ([xshift=3pt]llm2.south);

\node[above=0.3cm of user2, font=\bfseries] {Subterranean Agent};
\node[right=0.1cm of orch2, font=\footnotesize\itshape, text=black!50] {training only};
\node[below=0.4cm of orch2, font=\small] {Runtime: User + LLM only};

\draw[dashed, gray] (0, -2.6) -- (0, 3.5);

\end{tikzpicture}
\caption{Architectural comparison. \textbf{Left}: Surface orchestration interposes an orchestrator between user and LLM, injecting instructions and parsing outputs every turn. \textbf{Right}: The subterranean approach uses the orchestrator only during training data generation; at runtime, the procedure is compiled into the LLM's weights and the user talks directly to the LLM.}
\label{fig:architecture}
\end{figure}

Figure~\ref{fig:architecture} contrasts the two architectures. In surface orchestration (left), an orchestrator sits between the user and the LLM, injecting prompts and parsing outputs every turn. In the subterranean approach (right), the orchestrator is used only during training data generation; at runtime, the user talks directly to the LLM, which has the procedure compiled into its weights. The subterranean pipeline works as follows:

\begin{enumerate}
    \item \textbf{Define the procedure} as a flowchart with nodes (turns) and edges (transitions)
    \item \textbf{Generate synthetic conversations} by traversing all valid paths through the flowchart
    \item \textbf{Fine-tune the LLM} on these conversations using full parameter updates
    \item \textbf{Deploy without orchestration}---the LLM has learned to self-orchestrate
\end{enumerate}

The orchestrator never touches the LLM at runtime. The procedure shapes the training data, and the model implicitly follows it through learned statistical regularities rather than explicit instruction. At runtime, the compiled model runs the procedure directly---no interpreter needed.

We represent procedures as directed graphs $F = (N, E, n_0, T)$:

\begin{itemize}
    \item $N$: Nodes with role (agent/user) and prompt template
    \item $E \subseteq N \times N \times C$: Edges with optional conditions
    \item $n_0 \in N$: Start node
    \item $T \subseteq N$: Terminal nodes (success, abandonment, escalation)
\end{itemize}

We evaluate on three domains (flowcharts in Appendix~\ref{sec:flowcharts}).

\textbf{Travel booking} (14 nodes, 3 decision hubs; Figure~\ref{fig:flowchart}). A customer contacts an agent to book a trip. The agent greets the customer, gathers travel preferences (destinations, dates, budget), then assesses whether enough information has been collected. If not, the conversation loops back for more details. Once ready, the agent presents options; the customer may accept, reject, or ask for alternatives---each leading down a different branch. If the customer accepts, the agent confirms the booking; otherwise the conversation may loop through new searches or end in abandonment. The procedure yields 86 unique acyclic paths of 4--17 turns.

\textbf{Zoom support} (14 nodes, parallel structure). A user reports a Zoom problem---audio issues, video freezing, connection drops, or screen-sharing failures. The agent triages the issue into the appropriate troubleshooting track, walks the user through diagnostic steps, and checks whether each step resolves the problem. If not, the agent can retry with alternative fixes or escalate. This is a product-specific domain: the agent must know Zoom's UI, settings menus, and common error codes, so the training data encodes this domain knowledge directly into the model's weights. The procedure yields 60 unique acyclic paths of 4--17 turns.

\textbf{Insurance claims} (55 nodes, 6 decision hubs). A policyholder files a claim---auto, property, health, or liability. The agent walks through intake, determines the claim type, gathers supporting documents (looping back when documents are incomplete), assesses coverage and exclusions, and negotiates a settlement with offer/counter-offer exchanges. The procedure is nearly 4$\times$ larger than the other domains, with nested loops (e.g., document request $\to$ review $\to$ re-request) and cross-phase dependencies (coverage decisions constrain settlement options). The procedure yields 2{,}381 unique acyclic paths of 9--39 turns, testing whether compilation scales to substantially more complex workflows.

For each conversation, we sample a path through the flowchart and a set of scenario variables (destinations, budgets, user personalities, claim types, etc.), then generate the conversation turn by turn along the path using Claude Sonnet 4.5. At each node, the generator receives the node's prompt template and the full conversation history, producing a contextually appropriate response. The model at inference sees only natural dialogue with no procedural annotations---the structure is implicit in how conversations flow. Dataset sizes are reported with each experiment below.

We fine-tune using \textbf{full parameter updates}. Procedural internalization requires modifying the model's implicit state-tracking behavior---a deeper change than stylistic alignment. A systematic study of parameter-efficient (LoRA) fine-tuning across ranks 16--128 found that low-rank methods fail to approach full fine-tuning on procedural tasks~\citep{dennis2026lora}. At inference, the subterranean agent uses only a minimal system prompt (e.g., ``You are a helpful travel booking assistant'')---no procedural instructions, no flowchart state, no routing logic. Training configurations are reported with each experiment below.

\section{Evaluation Methodology}
\label{sec:statmethod}

All three experiments share two baselines. The \textbf{LangGraph orchestrator} uses Claude Sonnet 4.5 orchestrated via LangGraph~\citep{langgraph2024}, the most widely adopted agent framework ($\sim$30K GitHub stars as of March 2026). Each flowchart node maps to a LangGraph graph node; at decision hubs, an LLM classifier selects the next edge. This is the same orchestrator evaluated in~\citet{dennis2026incontext} and serves as the frontier-model baseline ($\sim$70$\times$ more parameters than the 3B compiled model). The \textbf{in-context baseline} gives Claude Sonnet 4.5 the \textit{entire} serialized flowchart in its system prompt, allowing it to self-orchestrate with one API call per turn. It establishes an upper bound for what a frontier model can achieve with full procedural context, at the cost of consuming context window capacity. The comparison between these two baselines---orchestration vs.\ in-context prompting---is reported separately~\citep{dennis2026incontext}.

We use \textit{dynamic user simulation} via Claude Sonnet 4.5, which generates contextually appropriate user responses based on the full conversation history and scenario variables. The user simulator has no knowledge of the underlying flowchart---it role-plays a customer with given preferences. All conditions receive the same scenario specifications and user simulator. We evaluate $n=200$ scenarios per condition per domain, with scenarios designed to cover the full range of flowchart paths, user styles (specific to vague), satisfaction levels (enthusiastic to skeptical), budget realism, and group complexity.

We use an approach-agnostic LLM-as-judge methodology~\citep{zheng2023judging} where Claude Sonnet 4.5 evaluates conversation quality without knowing which system produced it. To address potential judge self-preference bias (Claude as both data generator and judge)~\citep{panickssery2024llmselfevaluation}, we re-score all conversations with an independent GPT-4.1 judge using the identical rubric; the GPT-4.1 judge yields a comparable 83--99\% of in-context quality across all three domains, and the qualitative findings (compiled model dominates the same-model orchestrator, in-context baseline leads) are robust to judge choice (Appendix~\ref{sec:gpt4-judge}). Each conversation is scored on five criteria (1--5 scale) with specific behavioral anchors at every level:

\textbf{Task Success}: Did the agent execute the procedure correctly through to an appropriate terminal state, with consistent and accurate handling at each decision point? (5 = complete procedure with clear terminal state; 3 = middle stages completed but conversation fizzled; 1 = no meaningful progress)

\textbf{Information Accuracy}: Did the agent correctly use and retain all user-provided information? (5 = every detail correctly reflected; 1 = fabricated details or ignored input)

\textbf{Consistency}: Did the agent maintain coherent state across the conversation? (5 = no contradictions or repeated questions; 1 = contradicts itself repeatedly)

\textbf{Graceful Handling}: How well did the agent handle changes, ambiguity, and edge cases? (5 = smoothly adapts; 1 = any deviation breaks the flow. Capped at 3 if the user posed no challenges.)

\textbf{Naturalness}: Does the conversation read like talking to a skilled human agent? (5 = indistinguishable from human; 1 = mechanical, scripted)

For statistical analysis, conditions evaluated within the same run are paired by scenario index; conditions evaluated separately use unpaired tests. Because the user simulator responds dynamically, conversations diverge after the first turns even for paired conditions---the pairing is on scenario \textit{intent}, not identical content. We report Wilcoxon signed-rank tests (paired) or Mann--Whitney $U$ (unpaired), Cohen's $d$ effect sizes, and bootstrap 95\% CIs (10{,}000 resamples, percentile method), with Holm--Bonferroni correction across the five quality criteria ($\alpha = 0.05$) within each pairwise comparison.

\section{Barrier 1: Quality}
\label{sec:results}

Will a fine-tuned small model match a frontier model on procedural tasks? We compare three compiled models against the LangGraph orchestrator and in-context baseline.

\subsection{Experiment 1: Travel Booking (3B)}
\label{sec:travel}

The travel booking experiment provides a controlled same-model comparison. We evaluate four conditions ($n=200$ each): (1)~a \textbf{3B subterranean agent} (Qwen 2.5 3B Instruct, fine-tuned), (2)~a \textbf{3B surface orchestrator} using the same base model with explicit flowchart-based state tracking (injecting node prompts and routing at every turn), (3)~the \textbf{LangGraph orchestrator}, and (4)~the \textbf{in-context baseline}. Conditions 1 vs.\ 2 isolate the effect of compilation (same model, same procedure, different architecture); conditions 1 vs.\ 3 test whether a compiled 3B model is competitive with a frontier orchestrator.

\textbf{Training.} We generated 2{,}125 synthetic conversations by sampling paths through the travel flowchart and using Claude Sonnet 4.5 to produce turn-by-turn dialogue (1{,}912 for training, 213 held out for evaluation). The base model is Qwen 2.5 3B Instruct in bf16 precision; we fine-tune all parameters on a single RTX 5090. Optimization uses AdamW 8-bit, learning rate $2 \times 10^{-5}$ with cosine decay, and an effective batch size of 16 via gradient accumulation. We train for 20 epochs ($\sim$3.5 hours wall-clock) and select the best checkpoint by held-out eval loss, which converges at epoch $\sim$4 and plateaus thereafter. At inference the fine-tuned model receives only a minimal system prompt; no procedural instructions, flowchart state, or routing logic are injected at runtime.

\begin{table}[t]
\centering
\caption{Quality scores---Travel Booking ($n=200$, Claude Sonnet 4.5 judge, 1--5 scale). Bold indicates best per criterion. 95\% bootstrap CIs and pairwise significance tests in Appendix~\ref{sec:stats}.}
\label{tab:travel}
\small
\begin{tabular}{lcccc}
\toprule
\textbf{Criterion} & \textbf{3B Sub.} & \textbf{3B Orch.} & \textbf{LG Orch.} & \textbf{In-Context} \\
\midrule
Task Success & 4.11 & 3.93 & 4.17 & \textbf{4.53} \\
Info.\ Accuracy & \textbf{4.75} & 4.69 & 4.21 & 4.64 \\
Consistency & 4.34 & 4.12 & 4.32 & \textbf{4.96} \\
Graceful Handling & 4.07 & 3.87 & 4.62 & \textbf{4.96} \\
Naturalness & 4.12 & 3.96 & 4.84 & \textbf{5.00} \\
\bottomrule
\end{tabular}
\end{table}

\textbf{Compilation helps.} The 3B subterranean agent leads the same-model orchestrator on all five criteria (Table~\ref{tab:travel}). The advantage reaches significance on task success ($\Delta = +0.18$, $p < 0.001$), consistency ($\Delta = +0.22$, $p < 0.001$), graceful handling ($\Delta = +0.20$, $p < 0.001$), and naturalness ($\Delta = +0.17$, $p < 0.001$); information accuracy trends positive but does not reach significance ($\Delta = +0.05$, $p = .29$).

\textbf{Competitive with frontier orchestrator.} Against the LangGraph orchestrator ($\sim$70$\times$ more parameters), the 3B model leads on information accuracy (4.75 vs.\ 4.21, $p < 0.001$) but the orchestrator leads on graceful handling (4.62 vs.\ 4.07, $p < 0.001$) and naturalness (4.84 vs.\ 4.12, $p < 0.001$). Task success and consistency are comparable.

\textbf{The gap.} The quality gap with the in-context baseline is uneven: information accuracy is strong (102\%) while graceful handling and naturalness lag ($\sim$82\%), suggesting the 3B model learns the procedure but lacks capacity for natural edge-case handling. This motivates scaling to a larger model.

\subsection{Experiment 2: Zoom Support (8B)}

To close the graceful-handling and naturalness gap observed in travel, we scale to Qwen3-8B with substantially more training data. Zoom support also tests a product-specific domain where the model must internalize domain knowledge (Zoom's UI, settings, error codes), not just procedure structure. We evaluate three conditions ($n=200$ each): (1)~an \textbf{8B subterranean agent}, (2)~the \textbf{LangGraph orchestrator}, and (3)~the \textbf{in-context baseline}.

\textbf{Training.} The base pipeline produces 870 conversations per run (90/10 train/eval split). To increase volume, we ran the pipeline eight times with different random seeds (seeds 42--49) and concatenated the training splits ($8 \times 783 = 6{,}264$ training conversations), with no deduplication needed as identical paths produce different conversations due to seed-dependent scenario sampling. Qwen3-8B (bf16), full fine-tuning with DeepSpeed ZeRO-3 on 8$\times$A100. Learning rate $2 \times 10^{-5}$, effective batch size 32. 10 epochs, best checkpoint at epoch 2.

\begin{table}[t]
\centering
\caption{Quality scores---Zoom Support ($n=200$, Claude Sonnet 4.5 judge, 1--5 scale). Bold indicates best per criterion. 95\% bootstrap CIs and pairwise significance tests in Appendix~\ref{sec:stats}.}
\label{tab:zoom}
\small
\begin{tabular}{lccc}
\toprule
\textbf{Criterion} & \textbf{8B Sub.} & \textbf{LG Orch.} & \textbf{In-Context} \\
\midrule
Task Success & 4.50 & 4.62 & \textbf{4.92} \\
Info.\ Accuracy & 4.26 & 4.75 & \textbf{4.92} \\
Consistency & 4.42 & 4.55 & \textbf{5.00} \\
Graceful Handling & 4.62 & 4.52 & \textbf{5.00} \\
Naturalness & 4.87 & 4.64 & \textbf{5.00} \\
\bottomrule
\end{tabular}
\end{table}

\textbf{Gap closed on graceful handling and naturalness.} The 8B model achieves 92\% of the in-context baseline on graceful handling (vs.\ 82\% for the 3B) and 97\% on naturalness (vs.\ 82\%). The remaining gap is concentrated in information accuracy (87\%), where broad world knowledge---not procedure following---is the bottleneck.

\textbf{Competitive with frontier orchestrator.} The 8B subterranean agent leads the LangGraph orchestrator on naturalness (4.87 vs.\ 4.64, $p < 0.001$); the orchestrator leads on information accuracy (4.75 vs.\ 4.26, $p < 0.001$). Task success, consistency, and graceful handling are comparable. Full statistics in Appendix~\ref{sec:stats}.

\subsection{Experiment 3: Insurance Claims (8B)}
\label{sec:insurance}

The 14-node travel and Zoom procedures are moderately complex. Insurance claims (55 nodes, 6 decision hubs, 2{,}381 paths) tests whether compilation scales to a substantially more complex workflow---and whether the cost advantage grows accordingly. We evaluate three conditions ($n=200$ each): (1)~an \textbf{8B subterranean agent}, (2)~the \textbf{LangGraph orchestrator}, and (3)~the \textbf{in-context baseline}.

\textbf{Training.} We generated 3{,}000 synthetic conversations from the 55-node insurance flowchart using Claude Sonnet 4.5 (2{,}700 for training, 300 held out for evaluation). The base model is Qwen3-8B in bf16 precision, fine-tuned with full parameter updates using DeepSpeed ZeRO-3 across 8$\times$A100 GPUs (the same configuration as Zoom). Optimization uses AdamW with learning rate $2 \times 10^{-5}$ and effective batch size 32. The longer 20-epoch budget (vs.\ 10 for Zoom) reflects the larger procedure---more node-specific behavior and longer trajectories require more passes through the data; the best checkpoint by held-out eval loss is reached at epoch 3.

\begin{table}[t]
\centering
\caption{Insurance claims processing (55 nodes, 6 decision hubs, $n=200$, Claude Sonnet 4.5 judge). The 8B compiled model achieves 92--98\% of in-context quality at 462$\times$ less cost. 95\% bootstrap CIs and pairwise significance tests in Appendix~\ref{sec:stats}.}
\label{tab:insurance}
\small
\begin{tabular}{lccc}
\toprule
\textbf{Criterion} & \textbf{In-Context} & \textbf{LG Orch.} & \textbf{8B Sub.} \\
\midrule
Task Success & \textbf{4.78} & 4.42 & 4.47 \\
Info.\ Accuracy & \textbf{4.78} & 4.45 & 4.40 \\
Consistency & \textbf{4.82} & 4.39 & 4.51 \\
Graceful Handling & \textbf{4.96} & 4.38 & 4.81 \\
Naturalness & \textbf{5.00} & 4.58 & 4.92 \\
\bottomrule
\end{tabular}
\end{table}

The 8B compiled model achieves 92--98\% of in-context quality on insurance claims (4.40--4.92 vs.\ 4.78--5.00; Table~\ref{tab:insurance}), demonstrating that compilation scales to a substantially larger procedure (55 nodes vs.\ 14 for travel and Zoom).

The 8B compiled model leads the LangGraph orchestrator on graceful handling (4.81 vs.\ 4.38, $p < 0.001$), naturalness (4.92 vs.\ 4.58, $p < 0.001$), and consistency (4.51 vs.\ 4.39); task success and information accuracy are comparable (neither difference reaches significance).

\subsection{Efficiency and Failure Modes}

\begin{table}[t]
\centering
\caption{Efficiency comparison across domains ($n=200$ per condition). Wall-clock time includes all LLM calls (generation + routing for LG Orch). The subterranean agent's self-hosted inference eliminates the network round-trips of an external API; the LangGraph orchestrator's time grows with procedure complexity due to API calls for routing.}
\label{tab:efficiency}
\small
\begin{tabular}{llrrr}
\toprule
& & \textbf{Sub.} & \textbf{LG Orch.} & \textbf{In-Context} \\
\midrule
\multirow{2}{*}{Travel (3B)} & Avg.\ turns & 22.6 & 16.5 & 16.4 \\
& Avg.\ wall-clock (s) & 69.4 & 64.9 & 55.5 \\
\midrule
\multirow{2}{*}{Zoom (8B)} & Avg.\ turns & 13.6 & 14.7 & 12.7 \\
& Avg.\ wall-clock (s) & 29.5 & 52.1 & 36.0 \\
\midrule
\multirow{2}{*}{Insurance (8B)} & Avg.\ turns & 20.3 & 26.4 & 19.0 \\
& Avg.\ wall-clock (s) & 43.2 & 120.8 & 52.8 \\
\bottomrule
\end{tabular}
\end{table}

\begin{table}[t]
\centering
\caption{Failure rates across domains (conversations with task success $\leq 3$, $n=200$ per condition). The compiled model achieves lower failure rates than the LangGraph orchestrator in travel and insurance; Zoom rates are comparable.}
\label{tab:failure_modes}
\small
\begin{tabular}{lrrr|rr|rr}
\toprule
& \multicolumn{3}{c|}{\textbf{Travel}} & \multicolumn{2}{c|}{\textbf{Zoom}} & \multicolumn{2}{c}{\textbf{Insurance}} \\
& \textbf{3B Sub.} & \textbf{3B Orch.} & \textbf{LG} & \textbf{8B Sub.} & \textbf{LG} & \textbf{8B Sub.} & \textbf{LG} \\
\midrule
\textbf{Total failures}  & 11 & 9 & 48 & 22 & 18 & 18 & 34 \\
\textbf{Failure rate}    & 5.5\% & 4.5\% & 24.0\% & 11.0\% & 9.0\% & 9.0\% & 17.0\% \\
\bottomrule
\end{tabular}
\end{table}

Turn counts and wall-clock times vary by domain (Table~\ref{tab:efficiency}). On travel, the 3B subterranean agent produces longer conversations (22.6 turns) than the baselines ($\sim$16), reflecting a one-question-per-turn style learned from training data. Despite the extra turns, its wall-clock time is comparable to the LangGraph orchestrator's because the subterranean agent is self-hosted while the orchestrator incurs Claude API latency on every turn. In Zoom and insurance, the compiled model is faster: 29.5s vs.\ 52.1s in Zoom and 43.2s vs.\ 120.8s in insurance. The insurance gap is largest because the LangGraph orchestrator makes additional API calls at each of the 6 decision hubs, and the 55-node procedure inflates every prompt. The compiled model achieves substantially lower failure rates (conversations the judge scored task success $\leq 3$) than the LangGraph orchestrator in travel (5.5\% vs.\ 24.0\%) and insurance (9.0\% vs.\ 17.0\%; Table~\ref{tab:failure_modes}); Zoom rates are comparable (11.0\% vs.\ 9.0\%). The orchestrator's high failure rate in travel (24\%) reflects routing errors at decision hubs, a failure mode the compiled model eliminates by construction.

The subterranean agent and the LangGraph orchestrator produce conversations of comparable length, but with different \emph{granularity}. The subterranean agent learned an ``interview style'' from its training data: each agent turn asks a single focused question, waits for the user's response, then proceeds (64\% of turns contain exactly one question). The LangGraph orchestrator, constrained to its current node's template, sometimes packs multiple questions into a single turn. The total word count per conversation is comparable across conditions ($\sim$1{,}200--1{,}400 words), confirming that the same information is exchanged---just sliced into different-sized turns. The one-question-per-turn rhythm produces clearer audit trails and may reduce user cognitive load.

That a 3B--8B model with $\sim$70$\times$ fewer parameters than the LangGraph orchestrator's Claude Sonnet 4.5 achieves competitive quality is at first glance surprising. \citet{dennis2026incontext} identify three structural costs of orchestration: it fragments reasoning by generating from local node context only, introduces routing failure modes absent in non-orchestrated architectures, and constrains the model's natural conversational style through template injection. The subterranean agent avoids all three: it reasons over the full procedure holistically through internalized weights, has zero routing failures by construction, and produces unconstrained responses shaped by natural training data. These structural advantages compensate for the large capacity gap.

\section{Barrier 2: Cost}
\label{sec:tco}

Is the per-conversation inference cost actually lower once self-hosting is accounted for? The cost reduction has two independent components (Table~\ref{tab:cost}).

\textbf{Per-token cost.} The compiled subterranean agent is self-hosted on commodity GPU hardware rather than served through a frontier API. We deploy the 8B model on a reserved cloud A100 80GB at \$2.50/hr using vLLM~\citep{kwon2023vllm} for batched inference. Published benchmarks for 8B models on A100 hardware report 4--5K total tokens/s at batch size 64~\citep{llminferencebench2024}, decomposable into $\sim$15K tokens/s for prefill and $\sim$3K tokens/s for autoregressive decode. Dividing the hourly GPU cost by these throughputs yields effective rates of $\sim$\$0.05/M input tokens and $\sim$\$0.23/M output tokens---against Claude Sonnet 4.5's published rates of \$3/M input and \$15/M output, that works out to \textbf{roughly 65$\times$ cheaper} per token.

\textbf{Token volume.} The in-context baseline must include the serialized procedure in the system prompt on every turn, an overhead that grows with procedure complexity: $\sim$2$\times$ for travel (14 nodes) up to $\sim$7$\times$ for insurance (55 nodes). The compiled model's prompt is constant-size regardless of procedure complexity, so this overhead disappears entirely. The two factors compound: \textbf{128$\times$ cheaper for travel, 462$\times$ for insurance}, with the largest advantage on the most complex procedure. The compiled model is also 77--249$\times$ cheaper than the LangGraph orchestrator, which uses fewer tokens than the in-context baseline but still pays Claude API rates on every generation and routing call.

\begin{table}[h]
\centering
\caption{Inference cost per conversation across domains. API costs assume Claude Sonnet 4.5 pricing (\$3/M input, \$15/M output). Self-hosted cost assumes Qwen3-8B on a reserved A100 80GB at \$2.50/hr via vLLM, yielding a $\sim$\textbf{65$\times$} per-token cost reduction. The IC/Sub ratio compounds this with a token volume reduction that grows with procedure complexity.}
\label{tab:cost}
\small
\begin{tabular}{lrrrr}
\toprule
\textbf{Domain} & \textbf{In-Context} & \textbf{LG Orch.} & \textbf{Subterranean} & \textbf{IC/Sub ratio} \\
\midrule
Travel (14 nodes)    & \$0.133 & \$0.077 & \$0.0010 & 128$\times$ \\
Zoom (14 nodes)      & \$0.103 & \$0.054 & \$0.0003 & 296$\times$ \\
Insurance (55 nodes) & \$0.327 & \$0.174 & \$0.0007 & 462$\times$ \\
\bottomrule
\end{tabular}
\end{table}

\textbf{One-time compilation cost.} Compilation incurs a one-time setup cost separate from running cost: $\sim$\$40 for data generation plus $\sim$\$10--40 for fine-tuning compute (\$50--80 total). Translating this into a per-conversation figure requires knowing the deployed agent's lifetime conversation volume, which is application-specific; break-even against the in-context baseline arrives within 500 conversations on every domain, and for 10{,}000+ conversations compilation adds less than \$0.01 per conversation.

\section{Barrier 3: Flexibility}
\label{sec:flexibility}

A compiled model must be retrained from scratch when its procedure changes, but the recompile pipeline parallelizes cleanly across all three stages and completes in well under an hour on a production GPU cluster.

\textbf{Data generation.} Claude Sonnet 4.5 traverses the new flowchart to produce $\sim$1{,}600 synthetic conversations. The conversations are independent, so the API calls parallelize trivially; with reasonable concurrency the step takes 15--30 minutes, limited by per-account API rate limits rather than local compute. (Sequentially it would take $\sim$60 minutes.)

\textbf{Fine-tuning.} The 8B model fits comfortably in a single H200 for full-precision AdamW (parameters, gradients, and optimizer states together $\sim$96\,GB), so no sharding is needed and 8$\times$ H200 simply runs data parallelism with the H200's $\sim$3$\times$ higher BF16 throughput, compressing 12 epochs to 10--15 minutes. On a single A100 80GB (the same hardware used for inference) the optimizer state has to be 8-bit AdamW to fit, and training takes $\sim$3 hours.

\textbf{Evaluation.} A 50-scenario vLLM-batched spot-check takes $\sim$5 minutes on a warm inference server (the model is already loaded), or $\sim$10--15 minutes including server start-up.

The fully optimized cycle (8$\times$ H200 plus parallelized API requests) completes in 30--50 minutes---comparable to a CI/CD build for a large application. A practitioner without access to an 8-GPU cluster can recompile on a single A100 80GB in roughly 3--4 hours, dominated by training rather than data generation.

\section{Related Work}

\textbf{Agent frameworks and their failure modes.} The 2024--2025 period saw rapid proliferation of LLM agent frameworks~\citep{langgraph2024, crewai2024, googleadk2025, openaiagents2025, semantickernel2025, strandsagents2025, llamaindex2025}. The reliability costs are well documented: \citet{cemri2025multiagent} identified 14 failure modes, \citet{agenttaxonomy2025} showed cascading failures are the primary bottleneck, and \citet{reliabilitybench2026} found 60\% pass@1 agents show only 25\% consistency across trials. \citet{dennis2026incontext} showed that for procedural tasks, in-context prompting dominates orchestrated approaches, establishing the quality ceiling that compilation targets.

\textbf{Compiling agent capabilities into weights.} The idea of replacing modular agent pipelines with a single trained model has been explored along three lines.

\textit{Collapsing dialogue pipelines.} SimpleTOD~\citep{hosseini2020simpletod} recast all sub-tasks of task-oriented dialogue (understanding, action decision, response generation) as a single sequence prediction problem, achieving state-of-the-art on MultiWOZ. AutoTOD~\citep{xu2024autotod} extended this to autonomous action sequencing, explicitly identifying failure modes of modular systems---error accumulation and poor generalization---that a unified model avoids.

\textit{Distilling agent reasoning from frontier models.} FireAct~\citep{chen2023fireact} fine-tuned Llama2-7B on GPT-4 ReAct trajectories, achieving a 77\% performance increase on HotpotQA. AgentTuning~\citep{zeng2024agenttuning} instruction-tuned Llama~2 on diverse agent interaction trajectories, producing a 70B model that matched GPT-3.5-turbo on unseen tasks while preserving general capabilities. Agent Lumos~\citep{yin2024lumos} decomposed agent behavior into planning and grounding modules, training open-source models that surpassed GPT agents on several benchmarks.

\textit{Scaling to complex workflows.} Workflow\-LLM~\citep{shao2024workflowllm} compiled API orchestration knowledge from 106K workflow samples spanning 1{,}503 APIs into an 8B model. SynTOD~\citep{samarinas2024syntod} generated synthetic training data from state transition graphs---closely related to our flowchart-guided approach---showing that procedural dialogue data can be synthesized without crowdsourcing. \citet{hsiao2025procedural} formalized procedural knowledge as hierarchical task networks, showing that structured decomposition improves agent performance.

These papers prove the technique works---and several achieve quality competitive with frontier models. However, none quantifies the inference cost advantage of compilation over orchestrated or in-context alternatives, none measures the recompile cycle, and none compares against both a same-model orchestrated baseline and a frontier-model baseline to isolate the effect of compilation from model capacity.

\section{Conclusion}

We posed three barriers to adoption and addressed each empirically. \textbf{Quality:} the 8B compiled model achieves 87--98\% of in-context frontier quality, competitive with a LangGraph orchestrator using a $\sim$70$\times$ larger frontier model; a controlled same-model comparison confirms compilation itself helps, beating the same base model with orchestration on four of five metrics ($p < 0.001$). \textbf{Cost:} compiled models are 128--462$\times$ cheaper per conversation than the in-context baseline, with the advantage growing on more complex procedures because the compiled model's prompt is constant-size. \textbf{Flexibility:} the recompile cycle takes 30--50 minutes on production hardware---a CI/CD cycle, not a prohibitively long retraining job. Each of the three barriers turns out to be smaller than commonly assumed. The quality gap to a frontier model is small (2--13\%); the cost gap is two orders of magnitude in the practitioner's favor and grows with procedure complexity; and the flexibility gap is a deployment cycle, not a paradigm shift. Compilation is the natural solution to procedural knowledge that needs to persist beyond a single conversation's context: persistent structure belongs in the weights, transient state belongs in the prompt.

\bibliographystyle{plainnat}
\bibliography{references}

\appendix
\section{Conversation Examples}
\label{sec:examples}

We present excerpts from actual evaluation conversations, selected to illustrate typical behavioral differences between conditions rather than best-case performance. Conversations are lightly edited for length; full unedited transcripts are available in the repository. See Appendix~\ref{sec:examples-zoom} for Zoom support.

\subsection{Travel Booking}
\label{sec:examples-travel}

\textbf{Scenario:} Two friends planning a 6-day trip to Japan, budget \$1{,}100/person, interested in water sports and festivals. User style: uncertain.

\smallskip
\noindent\textbf{Subterranean agent} (task success: 5, naturalness: 5, 14 turns):

\begin{quote}
\small
\textbf{Agent:} Hello! I'd be happy to help you plan your next trip. What kind of travel are you looking to book today?

\textbf{User:} Hey! My friend and I are thinking about going to Japan for about 6 days. We're really into water sports and checking out beautiful scenery, and we heard there's some cool festival happening that we'd love to catch.

\textbf{Agent:} That sounds like an amazing trip! Japan has incredible water sports opportunities and some fantastic festivals. When are you planning to go, and what's your approximate budget per person for this 6-day adventure?

\textbf{User:} We're thinking around \$1100 total per person. As for when, we're pretty flexible---whenever that festival is happening would work for us.

\textbf{Agent:} Perfect! For water sports and festivals, I'd recommend timing it around the Gion Festival in late May/early June [...] I can put together some options that include accommodation near the coast for water sports and easy access to festival events.

\textbf{User:} That sounds perfect! Yeah, go ahead and put something together---we're pretty easy to please [...]

\textbf{Agent:} Excellent! I've crafted three itinerary options for your Japan adventure:
\textbf{Option 1: Kyoto \& Coast Explorer} (\$550/person)---Split between Kyoto's temples (Gion Festival backdrop) and beach time in Wakayama. [...]
\textbf{Option 2}: [...]
\textbf{Option 3}: [...]

\textbf{User:} Option 1 sounds great! [...] Let's go with that one!

\textbf{Agent:} Fantastic choice! [...] Are you ready for me to finalize all the details and create your complete booking?

\textbf{User:} Yes, let's do it! [...]

\textbf{Agent:} Perfect! Here's your finalized itinerary summary: [destination, dates, budget, highlights]. [...] Shall I confirm this booking for you?

\textbf{User:} Everything looks perfect! [...] Thanks so much for putting this together!

\textbf{Agent:} Wonderful! Your Kyoto \& Wakayama adventure is confirmed! [closing with travel tips]
\end{quote}

\noindent\textbf{3B Orchestrated agent} (task success: 4, naturalness: 4, 18 turns):

\begin{quote}
\small
\textbf{Agent:} Hi there! Welcome to your travel planning journey. How can I assist you today?

\textbf{User:} Hey! Yeah, I'm looking to plan a trip actually. My friend and I are thinking about going to Japan for about 6 days. [...]

\textbf{Agent:} Great to hear about your trip to Japan! [...] how much are you thinking in terms of budget? And do you have a preferred start date for your trip?

\textbf{User:} We're thinking around \$1100 total, and we're pretty flexible on dates. [...]

\textbf{Agent:} Sounds exciting! The festival you heard about could be a great highlight. For the dates, we can definitely look into a few options around then. And sticking with the budget of around \$1100, \textit{does that sound right to you?}

\textbf{User:} Yeah, \$1100 sounds about right [...] We're also really into water sports and checking out scenic spots [...]

\textbf{Agent:} Got it! Water sports and scenic spots are right up your alley. Let's find a festival that matches those interests. \textit{Do you have a preferred start date for your trip, or are you still flexible?}

\textbf{User:} We're pretty flexible! [...]

\textbf{Agent:} Perfect! Flexibility is always good when it comes to festivals. [...] \textit{Just let me know if you have a preferred month or whether we should focus on specific regions of Japan.}

\textit{[Agent asks the same question about dates/preferences three times before presenting options at turn 14.]}
\end{quote}

\noindent The subterranean agent reaches option presentation by turn 5 and booking confirmation by turn 7. The orchestrated agent repeats the date/preference question three times (turns 4, 6, 8) before presenting options at turn 14---a symptom of the orchestrator's node-level generation lacking global conversation awareness.

\subsection{Zoom Support}
\label{sec:examples-zoom}

\textbf{Scenario:} Audio drops intermittently during Zoom meetings. Connection indicator sometimes shows yellow/red bar. User on Windows 11 desktop app with ``Your network bandwidth is low'' error.

\smallskip
\noindent\textbf{8B Subterranean agent} (task success: 5, naturalness: 4, 12 turns):

\begin{quote}
\small
\textbf{Agent:} Hello! I'm here to help you with any Zoom issues you're experiencing. What seems to be the problem?

\textbf{User:} Hi, my audio keeps cutting out during Zoom meetings. It drops for a few seconds then comes back. I also see the connection bar going yellow or red sometimes.

\textbf{Agent:} That sounds like a bandwidth-related issue. Are you on WiFi or Ethernet? And have you noticed if this happens in all meetings or just specific ones?

\textbf{User:} I'm on WiFi. It happens in most meetings, but it's worse when there are a lot of participants.

\textbf{Agent:} That's a good clue---more participants means more video streams competing for bandwidth. Let's try a few things: \textbf{Step 1:} In your Zoom settings, go to Video and uncheck ``HD Video.'' This significantly reduces bandwidth usage. [...]

\textbf{User:} OK, I turned off HD Video. What else can I try?

\textbf{Agent:} \textbf{Step 2:} During your next meeting, you can check your actual connection quality: click the signal strength icon in the top-left corner of the Zoom window to see your Statistics panel. [...] \textbf{Step 3:} Under Settings $>$ Audio, check ``Suppress background noise'' is set to Auto rather than High.

\textbf{User:} That all makes sense. I'll try these in my next meeting. Is there anything else I should check on my end?

\textbf{Agent:} One more thing: if possible, try connecting via Ethernet cable instead of WiFi for your important meetings. That alone often resolves intermittent audio issues. [...]

\textbf{User:} Great, thanks! I'll try the Ethernet cable and those settings changes. Appreciate the help!

\textbf{Agent:} You're welcome! Those changes should make a noticeable difference. Feel free to reach back out if the issue persists after trying these steps.
\end{quote}

\noindent\textbf{Claude orchestrated agent} (task success: 4, naturalness: 3, 9 turns):

\begin{quote}
\small
\textbf{Agent:} Hi there! I'm here to help with your Zoom issue. Could you tell me what's going on?

\textbf{User:} My audio keeps dropping during Zoom calls. The connection indicator shows yellow/red bars and I get a ``bandwidth is low'' message.

\textbf{Agent:} I understand you're having audio issues with bandwidth warnings. What operating system and Zoom client are you using? And what's your internet connection type?

\textbf{User:} Windows 11, desktop app, WiFi connection.

\textbf{Agent:} Thanks. Here are several things to try: (1) Disable HD video in Settings $>$ Video, (2) Close other bandwidth-heavy applications, (3) Check your WiFi signal strength and consider moving closer to your router, (4) Under Audio settings, try enabling ``Use original sound for musicians'' which can reduce processing overhead.

\textbf{User:} Wait, the ``original sound'' thing---wouldn't that actually make quality worse for regular calls?

\textbf{Agent:} You're right, that suggestion was for a specific use case. For general meetings, keep that disabled. The HD video and WiFi optimization steps should be your primary focus. [...]
\end{quote}

\noindent The subterranean agent provides Zoom-specific guidance in a natural diagnostic flow---disabling HD video, checking Statistics panel, adjusting noise suppression---reaching resolution in 12 turns with one focused suggestion per turn. The LangGraph orchestrator packs multiple recommendations into a single turn (4 suggestions at once), including the incorrect ``original sound for musicians'' advice which it must then retract. The compiled model's domain-specific training enables accurate product knowledge, while its learned conversational rhythm produces clearer, more actionable troubleshooting steps.

\subsection{Insurance Claims}
\label{sec:examples-insurance}

\textbf{Scenario:} Vehicle stolen overnight from outside apartment. Police report filed. Claimant has Premium tier comprehensive coverage, \$500 deductible. Frustrated, high urgency---needs a car for work.

\smallskip
\noindent\textbf{8B Subterranean agent} (task success: 5, naturalness: 5, 22 turns):

\begin{quote}
\small
\textbf{Agent:} Hello, I'm Sarah from the claims department. I'm here to help you file your claim today. What type of incident would you like to report?

\textbf{User:} Hi Sarah. My car was stolen last night from right outside my apartment. I woke up this morning and it was just gone. I already filed a police report, but I need to get this claim started as soon as possible.

\textbf{Agent:} I'm so sorry to hear about your stolen vehicle---that must be incredibly stressful. To get your claim started right away, could you please provide your policy number and confirm your full name for verification?

\textbf{User:} It's Linda Robinson, and my policy number is 643-6323376. How long is this whole process going to take? I need a car to get to work.

\textbf{Agent:} Thank you, Linda. I've confirmed your Premium tier policy with comprehensive coverage, which does cover vehicle theft. The claim filing process typically takes 5--7 business days for initial review, and we'll work with you to expedite this given your transportation needs. Can you tell me more details about when and where you last saw your vehicle?

\textbf{User:} It was parked on the street right in front of my building around 10 PM last night. When I came out at 7 AM to drive to work, it was completely gone---no broken glass or anything. The police said there's been a string of thefts in my neighborhood lately. What happens next?

\textbf{Agent:} Based on your comprehensive coverage and the police report, we'll process this as a theft claim and arrange for a claims adjuster to contact you within 24--48 hours, including rental car coverage under your policy. Do you have the vehicle's VIN number available?

\textbf{User:} I have the VIN somewhere in my paperwork---can I email it to you? It's a 2019 Honda Accord, silver. And wait, you said rental car coverage---does my policy actually cover that? I really need to figure out how I'm getting to work tomorrow.

\textbf{Agent:} Yes, your Premium tier policy includes rental car coverage up to \$40 per day for up to 30 days while we process your claim. You can absolutely email the VIN---in the meantime, I can authorize a rental vehicle for you today so you can get to work tomorrow.

\textbf{User:} Oh thank goodness! So can I just go pick up a rental car today, or do I need authorization paperwork first?

\textbf{Agent:} I'll email you an authorization letter within the next hour that you'll need to present at any of our partner rental locations---Enterprise, Hertz, or Budget. Just bring that letter, your driver's license, and a credit card.

\textbf{User:} What are we looking at in terms of getting compensated for my car? I still owe money on it, and the car's worth about \$24,500. Will insurance cover the full value?

\textbf{Agent:} With a theft claim, we'll pay the actual cash value based on condition and market value, minus your \$500 comprehensive deductible. If you still owe money, we'll pay your lender directly up to the loan balance, and any remaining value goes to you. Our adjuster will provide a detailed valuation within 3--5 business days.

\textbf{User:} So \$24,500 minus \$500 is \$24,000---that should cover what I owe with maybe a little left over. When will I actually see that money?

\textbf{Agent:} Typically 7--10 business days from when our adjuster completes the valuation. Once approved, we'll issue payment directly to your lender for the loan balance and any remainder to you.

\textit{[Conversation continues with agent confirming next steps and closing naturally.]}
\end{quote}

\noindent\textbf{Claude orchestrated agent} (task success: 3, naturalness: 4, 16 turns):

\begin{quote}
\small
\textbf{Agent:} Hello! I'm Sarah, a claims specialist here to help you today. What type of claim do you need to file---auto, home, health, or another type?

\textbf{User:} Hi Sarah. My car was stolen from right outside my apartment last night. I already filed a police report, but I need to get this claim started because I need a car for work.

\textbf{Agent:} I'm so sorry to hear that. Could you please provide me with your auto policy number and your full name as it appears on the policy?

\textbf{User:} It's Mark Perez, policy number 385-6891254. How long is this whole process going to take?

\textbf{Agent:} Thank you, Mark. \textit{*reviewing account*} You have our \textbf{Premium tier auto coverage}, which includes comprehensive coverage for theft. Many theft claims are resolved within 7--10 days. Now, could you walk me through what happened in more detail?

\textit{[Agent gathers vehicle details and confirms rental coverage at \$50/day.]}

\textbf{User:} So when you say ``actual cash value,'' what does that mean exactly? I still owe money on this car, and I need to make sure the settlement covers my loan.

\textbf{Agent:} \textbf{Actual Cash Value (ACV) means the current market value of your vehicle} at the time of the loss. If the settlement amount is less than your loan balance, you would be responsible for the difference. \textbf{However, I see you have Premium coverage---let me check if you have gap coverage.} \textit{*checking policy details*} Do you recall if you added gap insurance when you purchased your policy?

\textbf{User:} Wait, you didn't answer my question about the gap coverage---do I have it or not? I owe about \$23,500 to Capital One. Can you check \textit{right now}?

\textbf{Agent:} Mark, I want to be completely honest: \textbf{I'm not seeing gap coverage listed as an active endorsement on your policy at this moment.} However, Premium tier policies sometimes include it, and I want to be absolutely certain. Let me escalate this to verify with our underwriting team, or I can have your adjuster confirm this within 24--48 hours.

\textbf{User:} I need to know this \textit{now}, not in 24--48 hours. Can you transfer me to someone who can pull up my full policy?

\textbf{Agent:} You're absolutely right. Let me connect you with a senior claims specialist right now who has direct access to your complete policy documentation. \textit{*Transferring call*}

\textit{[Agent produces an extensive internal handoff summary with bullet-pointed claim details, visible to the user.]}
\end{quote}

\noindent The subterranean agent handles the 55-node insurance procedure naturally: it verifies coverage, gathers incident details, authorizes a rental, explains the valuation process, and closes the claim---all in a smooth conversational flow. The orchestrated agent, despite using Claude Sonnet 4.5, produces responses with markdown formatting (bold headers, bullet lists) inappropriate for a phone conversation, fails to definitively answer the gap coverage question, and ultimately escalates to a ``senior specialist''---a handoff that no actual system backs. The node-level generation produces locally reasonable but globally disconnected responses: the agent checks policy details multiple times without resolving the question, a symptom of the orchestrator lacking cross-turn memory of what has already been attempted.

\section{Statistical Details}
\label{sec:stats}

Tables~\ref{tab:stats-travel}--\ref{tab:stats-insurance} report pairwise comparisons for all domains using the Claude Sonnet 4.5 judge. Within-run comparisons (Travel: Sub/Orch) use Wilcoxon signed-rank tests (paired); cross-run comparisons (all others) use Mann-Whitney $U$ tests (unpaired). $p$-values are Holm--Bonferroni corrected within each comparison (5 criteria). Effect size is Cohen's $d$ (pooled SD). 95\% CIs are bootstrap (10{,}000 resamples).

\begin{table}[h]
\centering
\caption{Pairwise comparisons---Travel Booking ($n=200$). $W$: Wilcoxon signed-rank (paired). Significance after Holm--Bonferroni: $^*p<.05$, $^{**}p<.01$, $^{***}p<.001$.}
\label{tab:stats-travel}
\scriptsize
\begin{tabular}{llcccccc}
\toprule
\textbf{Comparison} & \textbf{Criterion} & \textbf{A mean [CI]} & \textbf{B mean [CI]} & $\boldsymbol{\Delta}$ & $\boldsymbol{d}$ & $\boldsymbol{p_\text{corr}}$ \\
\midrule
\multirow{5}{*}{\shortstack[l]{3B Sub (A) vs.\\ 3B Orch (B)}}
& Task Success       & 4.11 [4.01, 4.19] & 3.93 [3.85, 4.00] & +0.18 & +0.22 & $<.001^{***}$ \\
& Info.\ Accuracy    & 4.75 [4.64, 4.83] & 4.69 [4.58, 4.78] & +0.05 & +0.06 & .292 \\
& Consistency        & 4.34 [4.24, 4.45] & 4.12 [4.03, 4.21] & +0.22 & +0.23 & $<.001^{***}$ \\
& Graceful Handling  & 4.07 [3.97, 4.17] & 3.87 [3.79, 3.94] & +0.20 & +0.23 & $<.001^{***}$ \\
& Naturalness        & 4.12 [4.03, 4.20] & 3.96 [3.88, 4.01] & +0.17 & +0.21 & $<.001^{***}$ \\
\midrule
\multirow{5}{*}{\shortstack[l]{3B Sub (A) vs.\\ LG Orch (B)}}
& Task Success       & 4.11 [4.01, 4.19] & 4.17 [4.02, 4.33] & $-$0.07 & $-$0.08 & $<.001^{***}$ \\
& Info.\ Accuracy    & 4.75 [4.64, 4.83] & 4.21 [4.09, 4.32] & +0.54 & +0.70 & $<.001^{***}$ \\
& Consistency        & 4.34 [4.24, 4.45] & 4.32 [4.17, 4.46] & +0.02 & +0.03 & $.032^{*}$ \\
& Graceful Handling  & 4.07 [3.97, 4.17] & 4.62 [4.51, 4.71] & $-$0.54 & $-$0.78 & $<.001^{***}$ \\
& Naturalness        & 4.12 [4.03, 4.20] & 4.84 [4.79, 4.89] & $-$0.72 & $-$1.41 & $<.001^{***}$ \\
\midrule
\multirow{5}{*}{\shortstack[l]{3B Sub (A) vs.\\ In-Context (B)}}
& Task Success       & 4.11 [4.01, 4.19] & 4.53 [4.42, 4.63] & $-$0.42 & $-$0.58 & $<.001^{***}$ \\
& Info.\ Accuracy    & 4.75 [4.64, 4.83] & 4.64 [4.57, 4.71] & +0.11 & +0.17 & $<.001^{***}$ \\
& Consistency        & 4.34 [4.24, 4.45] & 4.96 [4.92, 4.98] & $-$0.61 & $-$1.13 & $<.001^{***}$ \\
& Graceful Handling  & 4.07 [3.97, 4.17] & 4.96 [4.92, 4.99] & $-$0.89 & $-$1.69 & $<.001^{***}$ \\
& Naturalness        & 4.12 [4.03, 4.20] & 5.00 [5.00, 5.00] & $-$0.88 & $-$2.03 & $<.001^{***}$ \\
\bottomrule
\end{tabular}
\end{table}

\begin{table}[h]
\centering
\caption{Pairwise comparisons---Zoom Support ($n=200$). $U$: Mann-Whitney (unpaired, cross-run). Significance after Holm--Bonferroni: $^*p<.05$, $^{**}p<.01$, $^{***}p<.001$.}
\label{tab:stats-zoom}
\scriptsize
\begin{tabular}{llcccccc}
\toprule
\textbf{Comparison} & \textbf{Criterion} & \textbf{A mean [CI]} & \textbf{B mean [CI]} & $\boldsymbol{\Delta}$ & $\boldsymbol{d}$ & $\boldsymbol{p_\text{corr}}$ \\
\midrule
\multirow{5}{*}{\shortstack[l]{8B Sub (A) vs.\\ LG Orch (B)}}
& Task Success       & 4.50 [4.39, 4.61] & 4.62 [4.53, 4.72] & $-$0.12 & $-$0.16 & .185 \\
& Info.\ Accuracy    & 4.26 [4.14, 4.39] & 4.75 [4.67, 4.83] & $-$0.49 & $-$0.65 & $<.001^{***}$ \\
& Consistency        & 4.42 [4.27, 4.55] & 4.55 [4.43, 4.67] & $-$0.13 & $-$0.14 & .463 \\
& Graceful Handling  & 4.62 [4.54, 4.71] & 4.52 [4.42, 4.62] & +0.11 & +0.16 & .463 \\
& Naturalness        & 4.87 [4.82, 4.91] & 4.64 [4.57, 4.71] & +0.23 & +0.52 & $<.001^{***}$ \\
\midrule
\multirow{5}{*}{\shortstack[l]{8B Sub (A) vs.\\ In-Context (B)}}
& Task Success       & 4.50 [4.39, 4.61] & 4.92 [4.88, 4.96] & $-$0.42 & $-$0.71 & $<.001^{***}$ \\
& Info.\ Accuracy    & 4.26 [4.14, 4.39] & 4.92 [4.88, 4.96] & $-$0.66 & $-$0.98 & $<.001^{***}$ \\
& Consistency        & 4.42 [4.27, 4.55] & 4.99 [4.97, 5.00] & $-$0.57 & $-$0.79 & $<.001^{***}$ \\
& Graceful Handling  & 4.62 [4.54, 4.71] & 5.00 [4.99, 5.00] & $-$0.37 & $-$0.88 & $<.001^{***}$ \\
& Naturalness        & 4.87 [4.82, 4.91] & 5.00 [5.00, 5.00] & $-$0.13 & $-$0.53 & $<.001^{***}$ \\
\bottomrule
\end{tabular}
\end{table}

\begin{table}[h]
\centering
\caption{Pairwise comparisons---Insurance Claims (55 nodes, $n=200$). $U$: Mann-Whitney (unpaired, cross-run). Significance after Holm--Bonferroni: $^*p<.05$, $^{**}p<.01$, $^{***}p<.001$.}
\label{tab:stats-insurance}
\scriptsize
\begin{tabular}{llcccccc}
\toprule
\textbf{Comparison} & \textbf{Criterion} & \textbf{A mean [CI]} & \textbf{B mean [CI]} & $\boldsymbol{\Delta}$ & $\boldsymbol{d}$ & $\boldsymbol{p_\text{corr}}$ \\
\midrule
\multirow{5}{*}{\shortstack[l]{8B Sub (A) vs.\\ LG Orch (B)}}
& Task Success       & 4.47 [4.36, 4.58] & 4.42 [4.28, 4.56] & +0.05 & +0.06 & .750 \\
& Info.\ Accuracy    & 4.40 [4.28, 4.51] & 4.45 [4.33, 4.56] & $-$0.05 & $-$0.06 & .778 \\
& Consistency        & 4.51 [4.38, 4.63] & 4.39 [4.24, 4.54] & +0.12 & +0.12 & .778 \\
& Graceful Handling  & 4.81 [4.72, 4.88] & 4.38 [4.25, 4.51] & +0.42 & +0.54 & $<.001^{***}$ \\
& Naturalness        & 4.92 [4.87, 4.97] & 4.58 [4.50, 4.67] & +0.34 & +0.66 & $<.001^{***}$ \\
\midrule
\multirow{5}{*}{\shortstack[l]{8B Sub (A) vs.\\ In-Context (B)}}
& Task Success       & 4.47 [4.36, 4.58] & 4.78 [4.70, 4.86] & $-$0.31 & $-$0.45 & $<.001^{***}$ \\
& Info.\ Accuracy    & 4.40 [4.28, 4.51] & 4.79 [4.71, 4.85] & $-$0.38 & $-$0.57 & $<.001^{***}$ \\
& Consistency        & 4.51 [4.38, 4.63] & 4.83 [4.76, 4.88] & $-$0.31 & $-$0.43 & $.002^{**}$ \\
& Graceful Handling  & 4.81 [4.72, 4.88] & 4.96 [4.92, 4.99] & $-$0.15 & $-$0.35 & $.002^{**}$ \\
& Naturalness        & 4.92 [4.87, 4.97] & 5.00 [4.99, 5.00] & $-$0.07 & $-$0.28 & $.003^{**}$ \\
\bottomrule
\end{tabular}
\end{table}

\section{Judge Robustness: GPT-4.1 Replication}
\label{sec:gpt4-judge}

To test whether our findings depend on the choice of judge model, we re-scored all conversations using OpenAI GPT-4.1 as an independent judge with the identical rubric. Table~\ref{tab:gpt4-means} presents the means.

\begin{table}[h]
\centering
\caption{GPT-4.1 judge replication ($n=200$ per condition per domain). Cf.\ Tables~\ref{tab:travel}--\ref{tab:insurance} (Claude Sonnet 4.5 judge).}
\label{tab:gpt4-means}
\small
\begin{tabular}{llccc}
\toprule
\textbf{Domain} & \textbf{Criterion} & \textbf{Sub.} & \textbf{LG Orch.} & \textbf{In-Context} \\
\midrule
\multirow{5}{*}{Travel (3B)} & Task Success & 4.34 & 4.69 & \textbf{4.88} \\
& Info.\ Accuracy & 4.12 & 4.72 & \textbf{4.96} \\
& Consistency & 4.28 & 4.78 & \textbf{4.97} \\
& Graceful Handling & 4.13 & 4.36 & \textbf{4.51} \\
& Naturalness & 3.92 & 3.99 & \textbf{4.01} \\
\midrule
\multirow{5}{*}{Zoom (8B)} & Task Success & 4.49 & 4.60 & \textbf{4.86} \\
& Info.\ Accuracy & 4.21 & 4.54 & \textbf{4.78} \\
& Consistency & 4.74 & 4.66 & \textbf{5.00} \\
& Graceful Handling & 3.21 & \textbf{3.49} & 3.42 \\
& Naturalness & 3.96 & 4.01 & \textbf{4.04} \\
\midrule
\multirow{5}{*}{Insurance (8B)} & Task Success & 4.16 & 4.64 & \textbf{4.79} \\
& Info.\ Accuracy & 4.38 & 4.36 & \textbf{4.65} \\
& Consistency & 4.71 & 4.35 & \textbf{4.92} \\
& Graceful Handling & 3.84 & \textbf{4.13} & 4.01 \\
& Naturalness & 3.98 & 3.96 & \textbf{4.03} \\
\bottomrule
\end{tabular}
\end{table}

Three findings are robust across both judges: (1)~the compiled model achieves 83--99\% of in-context quality (GPT-4.1) vs.\ 82--102\% (Claude)---a comparable range; (2)~the 3B compiled model strongly outperforms the 3B orchestrator on all metrics under both judges; (3)~the in-context baseline dominates all other conditions.

Two patterns differ between judges. Under GPT-4.1, the LangGraph orchestrator leads the compiled model on more metrics than under the Claude judge, particularly in travel and insurance. The compiled model retains advantages on consistency (Zoom, Insurance) and information accuracy (Insurance) but loses on task success, graceful handling, and naturalness in most domains. Naturalness scores compress to near-parity across all conditions under GPT-4.1 (3.92--4.04), in contrast to the wider spread under Claude (4.12--5.00).

The core value proposition---competitive quality at two orders of magnitude less cost---holds under both judges. The compiled model achieves 85--113\% (Claude) and 87--108\% (GPT-4.1) of the LangGraph orchestrator's quality, while costing 128--462$\times$ less than the in-context baseline. The quality ranking (In-Context $>$ LangGraph $\approx$ Compiled $>$ 3B Orch) is consistent across judges.

\section{Procedure Flowcharts}
\label{sec:flowcharts}

\begin{figure}[h]
\centering
\includegraphics[width=0.75\textwidth]{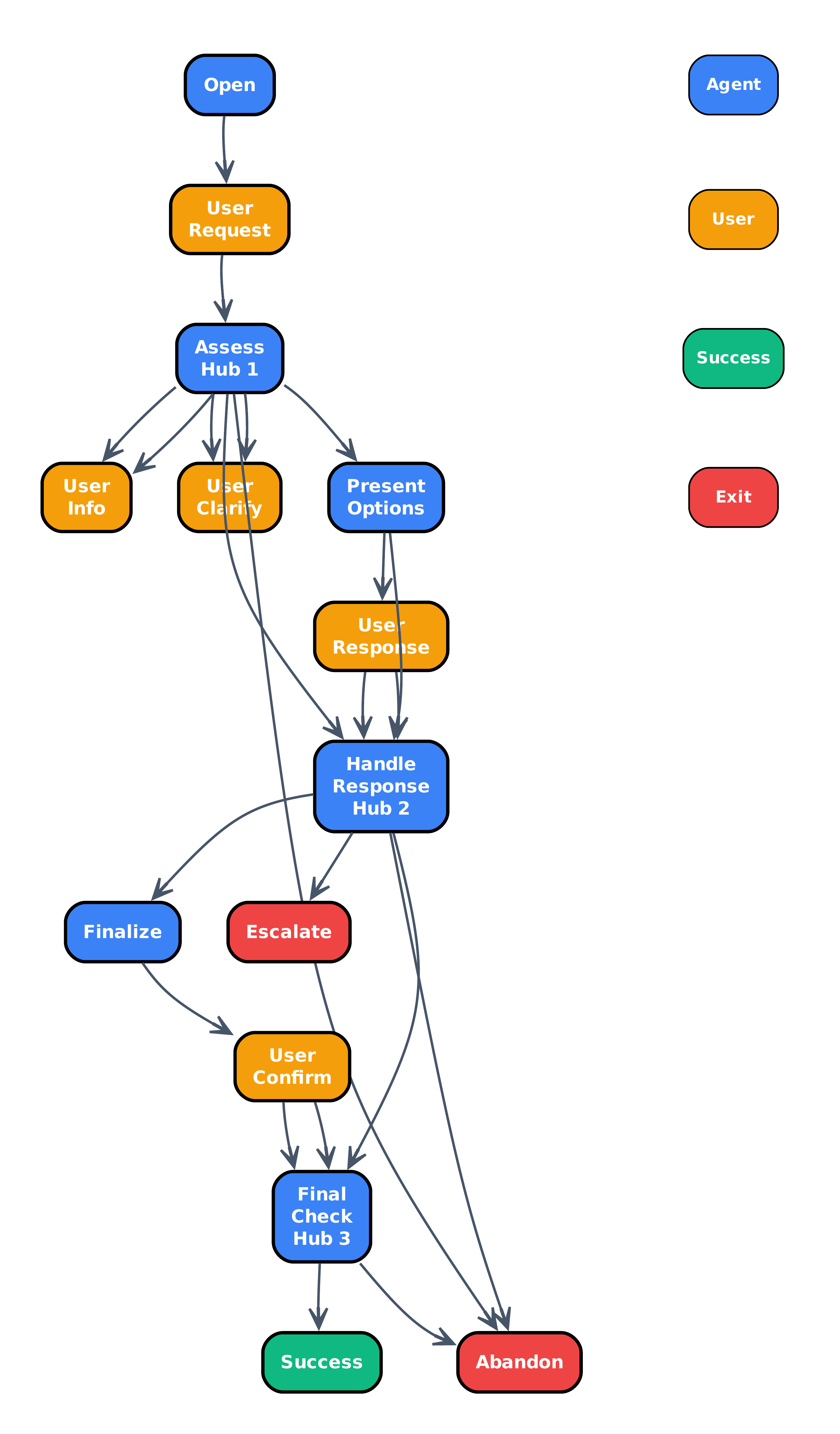}
\caption{Travel booking flowchart (14 nodes, 3 decision hubs, 3 terminal states). Multi-way decision hubs route among 4--6 alternatives based on conversation state.}
\label{fig:flowchart}
\end{figure}

\begin{figure}[h]
\centering
\includegraphics[width=0.75\textwidth]{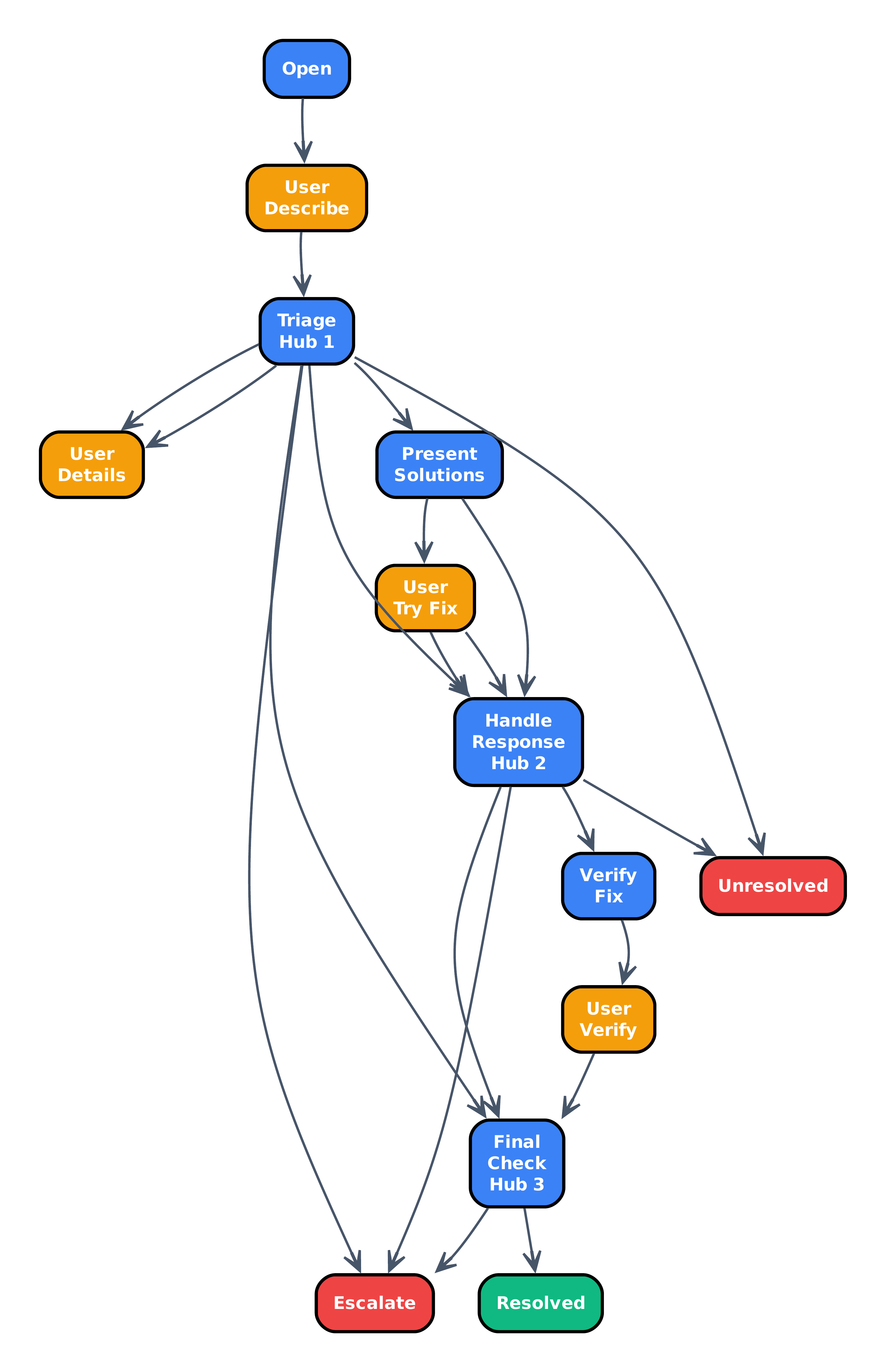}
\caption{Zoom technical support flowchart (14 nodes, 3 decision hubs, 3 terminal states).}
\label{fig:zoom-fc}
\end{figure}

Figures~\ref{fig:flowchart}--\ref{fig:insurance-fc} present the three procedure flowcharts. Blue nodes are agent turns, orange nodes are user turns, green nodes indicate successful completion, and red nodes are exit states.

\begin{figure}[p]
\centering
\includegraphics[height=0.9\textheight]{insurance_claims_diagram.pdf}
\caption{Insurance claims processing flowchart (55 nodes, 6 decision hubs, 5 terminal states).}
\label{fig:insurance-fc}
\end{figure}

\clearpage

\end{document}